\documentclass[12pt,reqno]{amsart}
\usepackage{amssymb}
\usepackage{amsmath, mathtools}

\usepackage{amsthm}

\usepackage{amscd}

\newcommand{\RNum}[1]{\uppercase\expandafter{\romannumeral #1\relax}}

\usepackage{caption}

\usepackage[T2A]{fontenc}
\usepackage[utf8]{inputenc}
\usepackage[english]{babel}

\input{int.def} 

\usepackage[sort]{cite}
\usepackage{tikz-cd}
\usetikzlibrary{cd}
\usepackage{dirtytalk}
\usepackage[linktoc=page, colorlinks, linkcolor=blue, citecolor=blue]{hyperref}

\usepackage{xcolor}
\usepackage{centernot}

\usepackage{enumitem}

\usepackage{pgfplots}
\usepackage{multicol}

\pgfplotsset{compat=1.17}



\numberwithin{equation}{section}

\DeclarePairedDelimiterX \ip[2]{\langle}{\rangle}{#1,#2}
\DeclarePairedDelimiterXPP \Prob[1]{\mathbb{P}}\{\}{}{ #1} 
\DeclarePairedDelimiterXPP \Probevent[1]{\mathbb{P}}(){}{#1} 

\usepackage[mathcal]{euscript}

\usepackage{titlesec}
\titleformat{\section}[runin]{\bfseries}{\thesection.}{3pt}{}[.]

\usepackage{geometry}
\newgeometry{vmargin={25mm}, hmargin={22mm,22mm}, footskip=10mm}   

\begin{document}
\title[Deep belief networks are exact
]{Deep belief networks are exact}

\author{Gleb Smirnov}
\address{
Mathematical Sciences Institute, 
Australian National University, Canberra, Australia}
\email{gleb.smirnov@anu.edu.au}


\begin{abstract}
We prove that every strictly positive probability distribution on \(\{-1,1\}^n\) is represented exactly by a sigmoid belief network with finite parameters. This answers a question of Sutskever and Hinton. The proof upgrades their probability-sharing approximation 
to exact representation using Brouwer's 
fixed-point theorem.
\end{abstract}

\maketitle
\setcounter{section}{0}

\section{Main results}\label{intro} 
A sigmoid belief network consists of binary random vectors
\[
V_k\in\{-1,1\}^{d_k},\qquad k=0,\ldots,N.
\]
The network is generated from \(V_N\) to \(V_0\): the top layer \(V_N\) is sampled first, 
and the remaining layers are generated
successively downwards. The visible layer is \(V_0\); 
its law is the probability distribution represented by the network.
\smallskip%

Let
\[
\sigma(t)=\frac{1}{1+e^{-t}}
\]
be the logistic function. The coordinates of \(V_N\) are independent, with
\[
\mathbb P\bigl(V_N(j)=1\bigr)=\sigma\bigl(b_N(j)\bigr)\quad 
\text{for some \(b_N\in\mathbb R^{d_N}\).}
\]
For \(k=0,\ldots,N-1\), set:
\[
U_k=W_kV_{k+1}+b_k,
\]
where
\[
W_k\in\mathbb R^{d_k\times d_{k+1}},
\qquad
b_k\in\mathbb R^{d_k}.
\]
Conditional on \(V_{k+1}\), the coordinates of \(V_k\) are independent, with
\[
\mathbb P\bigl(V_k(j)=1\mid V_{k+1}\bigr)
=
\sigma\bigl(U_k(j)\bigr),
\]
where \(V_k(j)\) and \(U_k(j)\) denote the \(j\)th coordinates.
\smallskip%

Sutskever and Hinton \cite{Suts-Hinton} proved that deep and narrow
sigmoid belief networks are universal approximators. More precisely,
every probability distribution on \(\{-1,1\}^n\) can be approximated
arbitrarily well by a sigmoid belief network of maximal layer width
\(n+1\) and depth
\[
3(2^n-1)+1.
\]
Their construction gives arbitrary approximation, but not exact
representation with finite parameters. Sutskever and Hinton asked whether
every strictly positive distribution can be represented exactly. We
answer this question affirmatively.
\begin{theorem}\label{main}
Every probability distribution \(\pi\) on \(\{-1,1\}^n\) such that
\[
\pi(x)>0,\qquad x\in\{-1,1\}^n,
\]
is exactly representable by a sigmoid belief network with finite
parameters. The network may be chosen with maximal layer width \(n+1\)
and depth \(3(2^n-1)+1\).
\end{theorem}
Strict positivity is necessary. 
For finite weights and biases, 
the top layer has full support and all transition probabilities are strictly positive; hence, every configuration of \(V_0\)
has positive probability.
\smallskip%

Our proof is topological and uses Brouwer's fixed point theorem. It shows
that suitable finite network parameters exist, but does not provide an
algorithm for finding them. 
It would be interesting to find a constructive proof.
\smallskip%

In \S\,\ref{sharing}, we recall the sharing construction of Sutskever and Hinton,
rederive their approximation theorem, and study the limiting behavior of the
resulting networks. In \S\,\ref{proof}, we use this analysis to prove Theorem~\ref{main}.
\smallskip%

Sigmoid belief networks go back to Neal \cite{Neal}.
Their representational properties were 
studied by Le Roux and Bengio \cite{LeRoux-Bengio-1}, who raised the question of universality
under restrictions on the width. Sutskever and Hinton
\cite{Suts-Hinton} answered this question affirmatively.
Subsequent work sharpened these approximation results;
see \cite{LeRoux-Bengio-2, Montufar-Ay, Merkh-Montufar}.
These results concern arbitrary approximation rather than exact representation. To the best of our knowledge, the question of exact representation 
remained open.

\section{Sharing}\label{sharing}
We recall the sharing construction of Sutskever and Hinton \cite{Suts-Hinton}, 
written in our $\{-1,1\}$ notation.
\smallskip%

Enumerate:
\[
\{-1,1\}^n=\{x_0,\ldots,x_M\},
\qquad M=2^n-1.
\]
Consider two vectors \(x_0,x_i\) and \(p\in[0,1]\).
A \emph{sharing} from \(x_0\) to \(x_i\) with parameter \(p\)
transfers a fraction \(p\) of the mass 
at \(x_0\) to \(x_i\). 
It transforms a distribution \(\mu\) into
\[
\mu'(x_0)=(1-p)\mu(x_0),\qquad
\mu'(x_i)=\mu(x_i)+p\mu(x_0),
\]
and leaves all other probabilities unchanged.
\smallskip%

Sharings generate every distribution on \(\{-1,1\}^n\). Start with a distribution concentrated at the single point \(x_0\). 
For \(i=1,\ldots,M\), share from 
\(x_0\) to \(x_i\) with parameter
\(p_i\). Suppose that the desired masses
\[
\pi_1,\ldots,\pi_{i-1}
\]
have already been placed 
at \(x_1,\ldots,x_{i-1}\). The mass remaining
at \(x_0\) is
\[
\pi_0+\pi_i+\cdots+\pi_M.
\]
Hence,
\[
p_i=
\frac{\pi_i}{\pi_0+\pi_i+\cdots+\pi_M}
\]
places precisely the mass \(\pi_i\) at 
\(x_i\). Proceeding for \(i=1,\ldots,M\), we obtain \(\pi\).
\smallskip%

In our application, \(\pi_i>0\) for all \(i\). 
Hence, \(p_i\in(0,1)\), and the above formula recovers each \(p_i\) from
\(\pi\). Thus, the construction gives a 
one-to-one map between the parameter
vectors
\[
(p_1,\ldots,p_M)\in(0,1)^M
\]
and the strictly positive probability distributions on \(\{x_0,\ldots,x_M\}\).
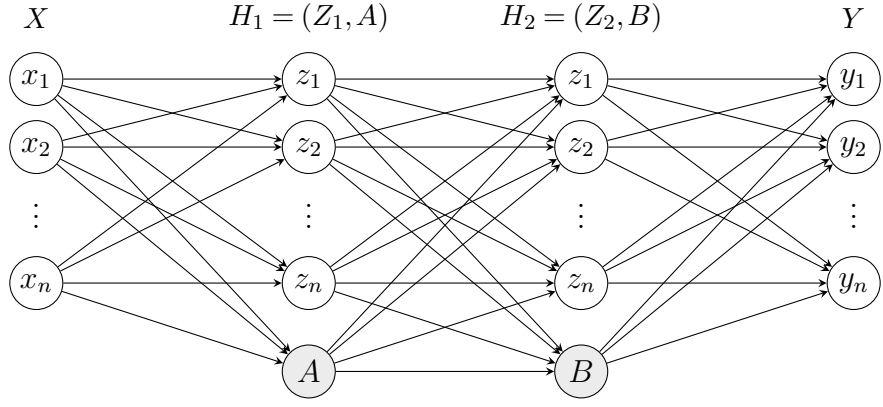
\begin{center}
\begin{tikzpicture}[
    x=1.8cm, y=0.9cm,
    >=stealth,
    neuron/.style={circle, draw, minimum size=7mm, inner sep=0pt},
    special/.style={circle, draw, minimum size=7mm, inner sep=0pt, fill=gray!15},
    lab/.style={font=\small}
]

\node[neuron] (x1) at (0,  1.5) {$x_1$};
\node[neuron] (x2) at (0,  0.5) {$x_2$};
\node            (x3) at (0, -0.4) {$\vdots$};
\node[neuron] (x4) at (0, -1.5) {$x_n$};
\node[lab] at (0, 2.4) {$X$};

\node[neuron]  (z11) at (2,  1.5) {$z_1$};
\node[neuron]  (z12) at (2,  0.5) {$z_2$};
\node          (z13) at (2, -0.4) {$\vdots$};
\node[neuron]  (z14) at (2, -1.5) {$z_n$};
\node[special] (A)   at (2, -2.8) {$A$};
\node[lab] at (2, 2.4) {$H_1=(Z_1,A)$};

\node[neuron]  (z21) at (4,  1.5) {$z_1$};
\node[neuron]  (z22) at (4,  0.5) {$z_2$};
\node          (z23) at (4, -0.4) {$\vdots$};
\node[neuron]  (z24) at (4, -1.5) {$z_n$};
\node[special] (B)   at (4, -2.8) {$B$};
\node[lab] at (4, 2.4) {$H_2=(Z_2,B)$};

\node[neuron] (y1) at (6,  1.5) {$y_1$};
\node[neuron] (y2) at (6,  0.5) {$y_2$};
\node         (y3) at (6, -0.4) {$\vdots$};
\node[neuron] (y4) at (6, -1.5) {$y_n$};
\node[lab] at (6, 2.4) {$Y$};

\foreach \u in {x1,x2,x4}
  \foreach \v in {z11,z12,z14,A}
    \draw[->, thin] (\u) -- (\v);

\foreach \u in {z11,z12,z14,A}
  \foreach \v in {z21,z22,z24,B}
    \draw[->, thin] (\u) -- (\v);

\foreach \u in {z21,z22,z24,B}
  \foreach \v in {y1,y2,y4}
    \draw[->, thin] (\u) -- (\v);

\end{tikzpicture}
\end{center}
We now recall how to approximate one sharing by a
sigmoid belief network. 
Fix \(i\in\{1,\ldots,M\}\) and \(p\in(0,1)\). The network has four layers
\[
\text{Input},\qquad H_1,\qquad H_2,\qquad \text{Output},
\]
of dimensions \(n,n+1,n+1,n\), 
respectively. Write
\[
X\in\{-1,1\}^n,\qquad
H_1=(Z_1,A),\qquad
H_2=(Z_2,B),\qquad
Y\in\{-1,1\}^n,
\]
where \(Z_1,Z_2\in\{-1,1\}^n\) and \(A,B\in\{-1,1\}\). Fix \(w>0\), and set:
\[
\varepsilon_w:=\sigma(-w)
=\frac{1}{1+e^w}
\le e^{-w}.
\]
\emph{Step 1: copy the input.}
For \(j=1,\ldots,n\), set:
\[
U_{Z_1}(j)=wX(j).
\]
Then:
\[
\mathbb P\bigl(Z_1(j)=X(j)\mid X\bigr)
=\sigma(w)=1-\varepsilon_w,
\]
and therefore:
\[
\mathbb P(Z_1\neq X\mid X)
\le n\varepsilon_w.
\]
\smallskip%

\emph{Step 2: detect \(x_0\).}
Set:
\[
U_A=w\bigl(\langle x_0,X\rangle-n+1\bigr).
\]
If \(X=x_0\), then \(U_A=w\), so:
\[
\mathbb P(A=-1\mid X=x_0)=\varepsilon_w.
\]
If \(X\neq x_0\), then:
\[
\langle x_0,X\rangle\le n-2,
\]
and hence \(U_A\le-w\). Thus:
\[
\mathbb P(A=1\mid X\neq x_0)\le\varepsilon_w.
\]
\smallskip%

\emph{Step 3: pass the input forward.}
For \(j=1,\ldots,n\), set:
\[
U_{Z_2}(j)=wZ_1(j).
\]
Then:
\[
\mathbb P(Z_2\neq Z_1\mid Z_1)
\le n\varepsilon_w.
\]
\smallskip%

\emph{Step 4: decide whether to share.}
Set:
\[
U_B=
\frac{w+\sigma^{-1}(p)}{2}\,A
+
\frac{\sigma^{-1}(p)-w}{2}.
\]
If \(A=1\), then:
\[
U_B=\sigma^{-1}(p),
\qquad
\mathbb P(B=1\mid A=1)=p.
\]
If \(A=-1\), then:
\[
U_B=-w,
\qquad
\mathbb P(B=1\mid A=-1)=\varepsilon_w.
\]
\smallskip%

\emph{Step 5: produce the output.}
For \(j=1,\ldots,n\), set:
\[
U_Y(j)=wZ_2(j)+w x_i(j)(B+1).
\]
If \(B=-1\), then:
\[
U_Y(j)=wZ_2(j),
\]
and hence:
\[
\mathbb P(Y\neq Z_2\mid Z_2,B=-1)
\le n\varepsilon_w.
\]
If \(B=1\), then:
\[
x_i(j)U_Y(j)
=
w x_i(j)Z_2(j)+2w
\ge w,
\]
and therefore:
\[
\mathbb P(Y\neq x_i\mid Z_2,B=1)
\le n\varepsilon_w.
\]
\smallskip%

Sharing is \emph{successful} if:
\begin{enumerate}[label=(\roman*)]
\item \(Z_1=X\);
\smallskip%

\item \(A\) correctly detects whether \(X=x_0\);
\smallskip%

\item \(Z_2=Z_1\);
\smallskip%

\item if \(A=-1\), then \(B=-1\);
\smallskip%

\item \(Y=Z_2\) if \(B=-1\), and \(Y=x_i\) if \(B=1\).
\end{enumerate}
If \(A=1\), no condition is imposed on \(B\); in this case,
\[
\mathbb P(B=1\mid A=1)=p
\]
is precisely the condition for the sharing.
\smallskip%

If successful, the network performs the sharing
exactly. By the union bound:
\[
\mathbb P(\text{sharing fails}\mid X)
\le
(3n+2)\varepsilon_w
\le
(3n+2)e^{-w}.
\]
Thus, as \(w\to\infty\), the network converges \emph{uniformly} in
\(p\in(0,1)\) and \(X\in\{-1,1\}^n\) to 
the sharing
\[
X\longmapsto
\begin{cases}
X, & X\neq x_0,\\[1mm]
x_i, & X=x_0 \text{ with probability }p,\\[1mm]
x_0, & X=x_0 \text{ with probability }1-p.
\end{cases}
\]
We now compose the sharings. Let
\[
p=(p_1,\ldots,p_M)\in(0,1)^M.
\]
Start with the distribution concentrated at \(x_0\).
For \(i=1,\ldots,M\), share from \(x_0\) to \(x_i\)
with parameter \(p_i\). Denote the resulting distribution by \(\Phi(p)\). Thus, \(\Phi\) is a one-to-one map from \((0,1)^M\) onto the
strictly positive probability distributions on
\(\{x_0,\ldots,x_M\}\).
\smallskip%

Replace each sharing by the sigmoid belief 
network above. The output layer of one sharing is the input layer of the next. Hence, starting with the initial layer, each sharing adds three more layers. Thus, \(M\) sharings is a 
network of depth
\[
3M+1=3(2^n-1)+1.
\]
Initialize the network. By definition, a sigmoid belief network
with finite parameters cannot have its top layer concentrated at a
single vector; therefore, choose
\[
b_N(j)=w x_0(j),\qquad j=1,\ldots,n.
\]
Then:
\[
\mathbb P\bigl(V_N(j)=x_0(j)\bigr)=\sigma(w),
\]
and hence:
\[
\mathbb P(V_N\neq x_0)\le n\varepsilon_w.
\]
Let \(\Phi_w(p)\) be the resulting 
distribution of \(V_0\). Initialization fails with probability at most \(n\varepsilon_w\),
and each of the \(M\) sharings fails with probability at most \((3n+2)\varepsilon_w\). Hence, the entire network can be coupled with the ideal sequence of sharings so that:
\[
\mathbb P(\text{the two outputs differ})
\le
\bigl[n+M(3n+2)\bigr]\varepsilon_w.
\]
Consequently:
\[
\|\Phi_w(p)-\Phi(p)\|_{\rm TV}
\le
\bigl[n+M(3n+2)\bigr]e^{-w}.
\]
In particular:
\[
\sup_{p\in(0,1)^M}
\|\Phi_w(p)-\Phi(p)\|_{\rm TV}
\longrightarrow0
\qquad\text{as }w\to\infty.
\]
The uniformity in \(p\) will be essential in passing from
approximation to exact representation.

\section{Proof of Theorem~\ref{main}}\label{proof} 
Let
\[
\Delta^\circ
=
\left\{
(\pi_0,\ldots,\pi_M):
\pi_i>0,\quad \sum_{i=0}^M\pi_i=1
\right\}
\]
be the interior of the probability simplex. 
From \S\,\ref{sharing}, the successive sharings define a continuous map
\[
\Phi:(0,1)^M\longrightarrow\Delta^\circ.
\]
Its inverse is given by
\[
\Phi^{-1}(\pi)_i
=
\frac{\pi_i}{\pi_0+\pi_i+\cdots+\pi_M},
\qquad i=1,\ldots,M.
\]
Thus, \(\Phi\) is a homeomorphism.
\smallskip%

For \(w>0\), let
\[
\Phi_w:(0,1)^M\longrightarrow\Delta^\circ
\]
be the map which assigns to \(p=(p_1,\ldots,p_M)\) the distribution
of the sigmoid belief network constructed in \S\,\ref{sharing}. The map
\(\Phi_w\) is continuous, and
\[
\sup_{p\in(0,1)^M}
\|\Phi_w(p)-\Phi(p)\|_{\rm TV}
\longrightarrow0
\qquad\text{as }w\to\infty.
\]
Fix the target distribution \(\pi\in\Delta^\circ\), 
and set:
\[
p^*:=\Phi^{-1}(\pi)\in(0,1)^M.
\]
Choose \(\delta>0\) so small that the closed cube
\[
D:=\{p\in\mathbb R^M:\|p-p^*\|_\infty\le\delta\}
\]
is contained in \((0,1)^M\). Since \(D\) is compact, \(\Phi(D)\subset\Delta^\circ\) is compact. Choose a compact neighborhood \(S\subset\Delta^\circ\) of \(\Phi(D)\). Since \(\Phi_w\to\Phi\) 
uniformly on \(D\), 
\[
\Phi_w(D)\subset S\quad \text{for all large enough \(w\)}.
\]
Since \(\Phi^{-1}\) is uniformly continuous on \(S\),
\[
\sup_{p\in D}
\left\|
\Phi^{-1}\bigl(\Phi_w(p)\bigr)-p
\right\|_\infty
\longrightarrow0.
\]
Choose \(w\) so large that
\[
\left\|
\Phi^{-1}\bigl(\Phi_w(p)\bigr)-p
\right\|_\infty
<\delta
\qquad\text{for all }p\in D.
\]
Define:
\[
T:D\longrightarrow \mathbb R^M,
\qquad
T(p):=
p^*+p-\Phi^{-1}\bigl(\Phi_w(p)\bigr).
\]
Then:
\[
\|T(p)-p^*\|_\infty
=
\left\|
p-\Phi^{-1}\bigl(\Phi_w(p)\bigr)
\right\|_\infty
<\delta.
\]
Hence, \(T(D)\subset D\). Moreover, 
\(T\) is continuous. By Brouwer's fixed point theorem, \(T\) has a fixed point
\(p\in D\):
\[
p
=
p^*+p-\Phi^{-1}\bigl(\Phi_w(p)\bigr).
\]
Hence:
\[
\Phi^{-1}\bigl(\Phi_w(p)\bigr)= p^* =\Phi^{-1}(\pi).
\]
Consequently, \(\Phi_w(p)=\pi\), and the proof follows. \qed

\bibliographystyle{plain}
\bibliography{ref}

\end{document}